\PassOptionsToPackage{numbers,sort&compress,square}{natbib}

\documentclass{article} 
\usepackage[preprint]{neurips_2025}

\usepackage[breaklinks=true]{hyperref}
\usepackage{url}
\usepackage{graphicx}
\usepackage{booktabs}
\usepackage{amssymb}
\usepackage{amsmath}
\usepackage{multirow}
\usepackage{multicol}

\usepackage{subcaption}

\usepackage{makecell}
\usepackage{pgffor}

\usepackage{graphicx}
\usepackage{booktabs}
\usepackage{amssymb}
\usepackage{amsmath}
\usepackage{multirow}
\usepackage{multicol}
\usepackage{subcaption}
\usepackage{makecell}
\usepackage{pgffor}
\usepackage[T1]{fontenc}
\usepackage{siunitx}
\usepackage{needspace}
\usepackage{caption}
\usepackage{placeins}
\usepackage{textcomp}
\usepackage{bm}
\usepackage{mdframed}
\newmdenv[
  linecolor=black,
  linewidth=0.5pt,
  roundcorner=4pt,
  skipabove=6pt,
  skipbelow=6pt
]{examplebox}

  {\begin{examplebox}\textbf{#1.} }%
  {\end{examplebox}}

\title{Curi\'o-Edu 7B: Examining Data Selection Impacts \\[0.1cm] in LLM Continued Pretraining}

\author{%
Thales Sales Almeida \\
  Institute of Computing (IC)\\
  University of Campinas (UNICAMP) \\
  Maritaca AI\\
  \And
Rodrigo Nogueira\\
  School of Electrical and Computer Engineering (FEEC)\\
University of Campinas (UNICAMP) \\
  Maritaca AI \\
  \And
Hélio Pedrini \\
  Institute of Computing (IC)\\
  University of Campinas (UNICAMP) \\
}

\begin{document}

\maketitle

\begin{abstract}
Continued pretraining extends a language model's capabilities by further exposing it to additional data, often tailored to a specific linguistic or domain context. This strategy has emerged as an efficient alternative to full retraining when adapting general-purpose models to new settings. In this work, we investigate this paradigm through Curió~7B, a 7-billion-parameter model derived from LLaMA-2 and trained on 100 billion Portuguese tokens from the ClassiCC-PT corpus — the most extensive Portuguese-specific continued-pretraining effort above the three-billion-parameter scale to date. Beyond scale, we investigate whether quantity alone suffices or whether data quality plays a decisive role in linguistic adaptation. To this end, we introduce Curió-Edu~7B, a variant trained exclusively on the educational and STEM-filtered subset of the same corpus, totaling just 10 billion tokens. Despite using only 10\% of the data and 20\% of the computation, Curió-Edu~7B surpasses the full-corpus model in our evaluations, demonstrating that data selection can be fundamental even when adapting models with limited prior exposure to the target language. The developed models are available at \url{https://huggingface.co/collections/ClassiCC-Corpus/curio-edu}
\end{abstract}

\section{Introduction}

Large Language Models (LLMs) have shown remarkable capabilities across a wide range of natural language understanding and generation tasks~\cite{application_1,santos2025bluex,application_2}. However, their performance remains uneven across languages, with high-resource languages such as English benefiting disproportionately from large-scale pretraining.

In this context, Portuguese is often considered a language with limited resources and comparatively little research attention~\cite{longpre2024bridgingprovenancegap}, despite being the fifth most spoken language worldwide. Addressing this gap is essential for expanding access to advanced language technologies in Portuguese-speaking regions and ensuring that progress in LLMs benefits a broader linguistic community.

Recent studies have demonstrated that continued pretraining--further training an existing model on specifics domain or languages—can substantially improve performance in underrepresented languages without the cost of training a model from scratch~\cite{sabia, almeida2025poetav2robustevaluation, curio1}. Yet the scale, methodology, and evaluation of continued pretraining for Portuguese remain underexplored.

Prior work has typically focused on smaller or instruction-tuned models, offering limited insight into how larger models adapt to new linguistic distributions and how data quality influences this adaptation~\cite{curio1,paiola2025bode2}. A fundamental question remains: is it more effective to expose a model with limited prior experience in a language to a small but high-quality corpus, or to a much larger corpus with lower average quality?

In this work, we introduce Curió~7B and Curió-Edu~7B, two 7-billion-parameter models derived from LLaMA-2~\cite{touvron2023llama2} and continued-pretrained exclusively in Portuguese. Using the ClassiCC-PT corpus~\cite{curio1}, we train Curió~7B on 100 billion unique tokens without semantic filtering, representing the most extensive Portuguese-focused continued-pretraining effort to date among models above 3B parameters.

For Curió-Edu~7B, we use the educational portion of ClassiCC-PT, totaling 20 billion tokens tagged by ClassiCC classifiers as having notable educational value. Starting from a base model with minimal Portuguese exposure enables us to isolate the effects of continued pretraining in the target language and to assess the influence of semantic data selection.

To assess the impact of these training regimes, we evaluate both models using the PoETa~V2~\cite{almeida2025poetav2robustevaluation} benchmark, a comprehensive suite of tasks covering linguistic, reasoning, and cultural dimensions of Portuguese.

Our analysis investigates how specialization affects accuracy, cross-domain generalization, and alignment with Portuguese linguistic norms. Through this study, we aim to shed light on the effectiveness of large-scale continued pretraining as a strategy for improving LLMs in low- to medium-resource languages and to provide a foundation for future model development.

\section{Related Work}

Research on improving language model performance across diverse linguistic and regional contexts has expanded significantly in recent years. Two threads are particularly relevant to our study: (i) analyses of geographic and cultural disparities in large language models, which highlight systemic gaps in representation and performance, and (ii) efforts to develop or adapt LLMs specifically for Portuguese, a language that remains comparatively under-resourced despite its global prevalence. The following subsections review these lines of work, outlining both the challenges posed by regional data scarcity and the current landscape of Portuguese-focused model development.

\subsection{LLM Gaps Based on Regional Context}

A growing body of research shows that large language models (LLMs) exhibit not only linguistic but also regional performance disparities. Beyond differences between high- and low-resource languages~\cite{low_resource_1,low_resource_2}, models often underperform on tasks requiring localized knowledge or cultural familiarity~\cite{moayeri2024worldbench,almeida2025tiebe,gao2025localbench}.

Recent benchmarks highlight this issue across several dimensions: everyday cultural reasoning~\cite{blend}, recall of socioeconomic and geographic indicators~\cite{moayeri2024worldbench}, and retrieval of regional or temporally grounded factual events~\cite{almeida2025tiebe}. Together, these findings reveal a persistent imbalance in how global models represent and reproduce regional knowledge.

A central cause lies in the uneven distribution of data used for pretraining. Analyses of publicly available resources, such as datasets hosted on Hugging Face, indicate that South America contributes less than 2\% of all benchmarks and datasets, with this share decreasing over time~\cite{longpre2024bridgingprovenancegap}. This distribution reflects broader structural inequalities in research production and digital data availability. Consequently, even multilingual models tend to inherit global English-centric biases, further amplifying regional knowledge gaps.

Continued pretraining has emerged as a practical strategy for mitigating these disparities. By adapting existing LLMs to language- or region-specific corpora, it enables improved coverage of underrepresented contexts without the cost of training a model from scratch. Understanding how data quality, topical focus, and scale interact in this process can offer valuable insights into how regional representation can be strengthened more efficiently.

\subsection{LLMs Specialized for Portuguese}

Efforts to improve LLM performance in Portuguese have advanced along multiple directions, ranging from encoder-based language models to large decoder-only architectures. Early initiatives such as BERTimbau~\cite{bertimbau}, PTT5~\cite{ptt5v2}, and Albertina~\cite{albertina} focused on pretraining from scratch or adapting masked language models to Portuguese. More recent work has shifted toward adapting large-scale generative models through continued pretraining and instruction tuning~\cite{paiola2025bode2, gervasio}.

A prominent trend involves extending multilingual or English-dominant base models with targeted Portuguese corpora. For example, Sabiá~\cite{sabia} adapts LLaMA~\cite{zhang2024tinyllama} 7B and 65B models using roughly 10B Portuguese tokens, showing that continued pretraining can substantially improve performance even when the original model had limited exposure to the language. Likewise, Curió~1.1B~\cite{curio1} builds on TinyLlama~\cite{zhang2024tinyllama} using 100B Portuguese tokens from the ClassiCC-PT corpus, achieving notable gains in comprehension and generation quality.

Another approach centers on monolingual pretraining and tokenizer adaptation. Cabrita~\cite{cabrita}, for instance, continually pretrains OpenLLaMA~3B, retraining its embeddings and optimizing its tokenizer for Portuguese, using approximately 7B tokens from mC4-PT~\cite{mc4}. Glória~\cite{gloria} trains GPT-Neo models (1.3B and 2.7B parameters) on 35B curated European Portuguese tokens, emphasizing literary and formal text. Other efforts, such as TeenyTinyLlama~\cite{correa2024teenytinyllama}, explore smaller-scale models (160M--460M) trained from scratch on 6B Portuguese tokens, while Tucano~\cite{correa2024tucano} represents a rare large monolingual initiative, pretraining models of up to 2.4B parameters exclusively on 500B Portuguese tokens from GigaVerbo.

In this work, we build upon these developments by introducing Curió~7B and Curió-Edu~7B, two LLaMA-2~\cite{touvron2023llama2}–based models continually pretrained exclusively on Portuguese data. Both rely on the ClassiCC-PT~\cite{curio1} corpus as the foundation for adaptation, enabling a controlled study of how different data selection strategies influence performance in a model with minimal prior exposure to Portuguese.

Curió~7B is trained on 100B unfiltered tokens, representing the largest Portuguese-specific continued-pretraining experiment above 3B parameters to date. In contrast, Curió-Edu~7B uses a 10B-token subset filtered for educational and STEM-related content via ClassiCC classifiers. Comparing these models allows us to disentangle the effects of data quantity versus semantic quality in continued pretraining for underrepresented languages.

\section{Methodology}
\label{sec:methodology}

This section describes the methodological framework adopted to investigate the effects of continued pretraining on Portuguese language models. We begin by detailing the construction and characteristics of the ClassiCC-PT corpus, including the processes used to derive both the full dataset and the educationally filtered subset. We then outline the training setup for all Curió~7B variants, highlighting architectural choices, computational budget, and differences between full-corpus and filtered-corpus pretraining. Finally, we present the evaluation protocol, including the PoETa~V2 benchmark and the metrics used to assess linguistic and task-oriented performance.

\subsection{Model Architecture and Training Setup}

Both Curió~7B and Curió-Edu~7B are based on the LLaMA 2~7B architecture. This choice was motivated by two main factors: (i) the public availability of its model weights and implementation, which facilitates reproducibility, and (ii) the transparency of its pretraining data composition.

According to the official LLaMA 2 documentation, Portuguese accounts for only 0.05\% of the total pretraining corpus—approximately 10~billion tokens. This makes LLaMA 2~7B a suitable base for studying linguistic adaptation under conditions of minimal prior exposure to Portuguese.

Both models were trained under a continued-pretraining regime, initializing from the original LLaMA 2~7B checkpoint. The standard Curió~7B model was trained for 100~billion tokens, while the educational variant Curió-Edu~7B was trained for 20~billion tokens. We adopted a sequence length of 4,096 tokens and applied sequence packing to maximize context utilization.

Optimization was performed using the Adafactor~\cite{shazeer2018adafactor} optimizer with a peak learning rate of $10^{-3}$ and a cosine decay schedule. We used a global batch size of 256 and employed mixed-precision training to reduce memory overhead and improve throughput.

Training was conducted on a TPU~v2-256 system using the T5x framework~\cite{roberts2023t5x}, which provides integrated support for both data and tensor parallelism. This setup enabled efficient scaling across TPU pods while maintaining stable performance. The estimated compute cost for training Curió~7B and Curió-Edu~7B is approximately $7{,}000$ and $1{,}400$, respectively\footnote{Training costs are estimated using TPU~v6 pricing, as TPU~v2 hardware is no longer commercially available.}.

In addition, we trained a smaller variant, Curió-Edu~1.1B, following the same training recipe as Curió-Edu~7B. This model provides a controlled comparison across two scales, allowing us to isolate the effects of dataset curation and data quality. For this analysis, we use the Curió 1.1B baseline introduced in prior works~\cite{curio1}.

\subsection{Pretraining Corpus}

All models in the Curió~7B family were pretrained exclusively on ClassiCC-PT, a 120-billion-token Portuguese corpus derived from filtered and processed Common Crawl snapshots. The dataset underwent extensive cleaning, deduplication, and quality scoring, producing a large-scale yet curated resource representative of a broad range of Portuguese-speaking domains. Due to computational budget constraints, we used 100 billion tokens from this corpus to train Curió~7B.

A distinctive characteristic of ClassiCC-PT is the inclusion of classifier-based document scores that estimate how strongly each document relates to educational or STEM content.

To analyze the role of data quality and topical focus in continued pretraining, we constructed a filtered subset by selecting only documents with an educational or STEM score above 2.5. This filtering step yielded approximately 10 billion tokens, which we used to train Curió-Edu~7B and Curió-Edu~1.1B. These variants were designed to evaluate whether a smaller, semantically filtered, higher-quality corpus can rival—or surpass—a much larger unfiltered dataset. Because the filtered subset was substantially smaller, we trained the educational variants for two full epochs, totaling 20 billion tokens seen during pretraining.

\subsection{Evaluation}

We evaluated Curió~7B and Curió-Edu~7B using PoETa~V2, a large-scale benchmark comprising more than 40 tasks designed to assess Portuguese language models across diverse domains and task types. PoETa~V2 covers a broad spectrum of areas—including reasoning, classification, mathematics, common sense, ethics, exams, and general knowledge—providing a comprehensive view of both linguistic and cognitive capabilities.

Our analysis examines both the overall performance across all PoETa~V2 tasks and the results within individual domains. This dual perspective allows us to investigate whether the semantic filtering applied to the Curió-Edu~7B training data yields measurable advantages in specific categories. Since the filtering procedure prioritizes educational and STEM-related content, our initial hypothesis was that improvements would be concentrated in the Math and Exams domains. In practice, however, we observed consistent performance gains across a wide range of areas.

All results are reported using the Normalized Preferred Metric (NPM), as defined in the PoETa~V2 paper~\cite{almeida2025poetav2robustevaluation}. NPM provides a unified scoring framework that normalizes heterogeneous task metrics, enabling consistent comparisons across models and categories. This evaluation setup allows for a direct examination of how data quality and corpus composition influence the outcomes of Portuguese-specific continued pretraining.





\section{Results}

This section presents the empirical findings of our continued pretraining experiments across multiple model sizes, data configurations, and evaluation domains. We begin by examining the overall progression of NPM throughout training, highlighting how different pretraining corpora impact learning dynamics. We then analyze the comparative performance of models trained on full versus semantically filtered datasets, followed by a detailed evaluation of the gains achieved by the Curió-Edu variants relative to their respective base models.

\subsection{Overall Performance Progression During Training}
\label{subsec:overall_performance_progression}

\begin{figure}[!htb]
\centering
\includegraphics[width=1\linewidth]{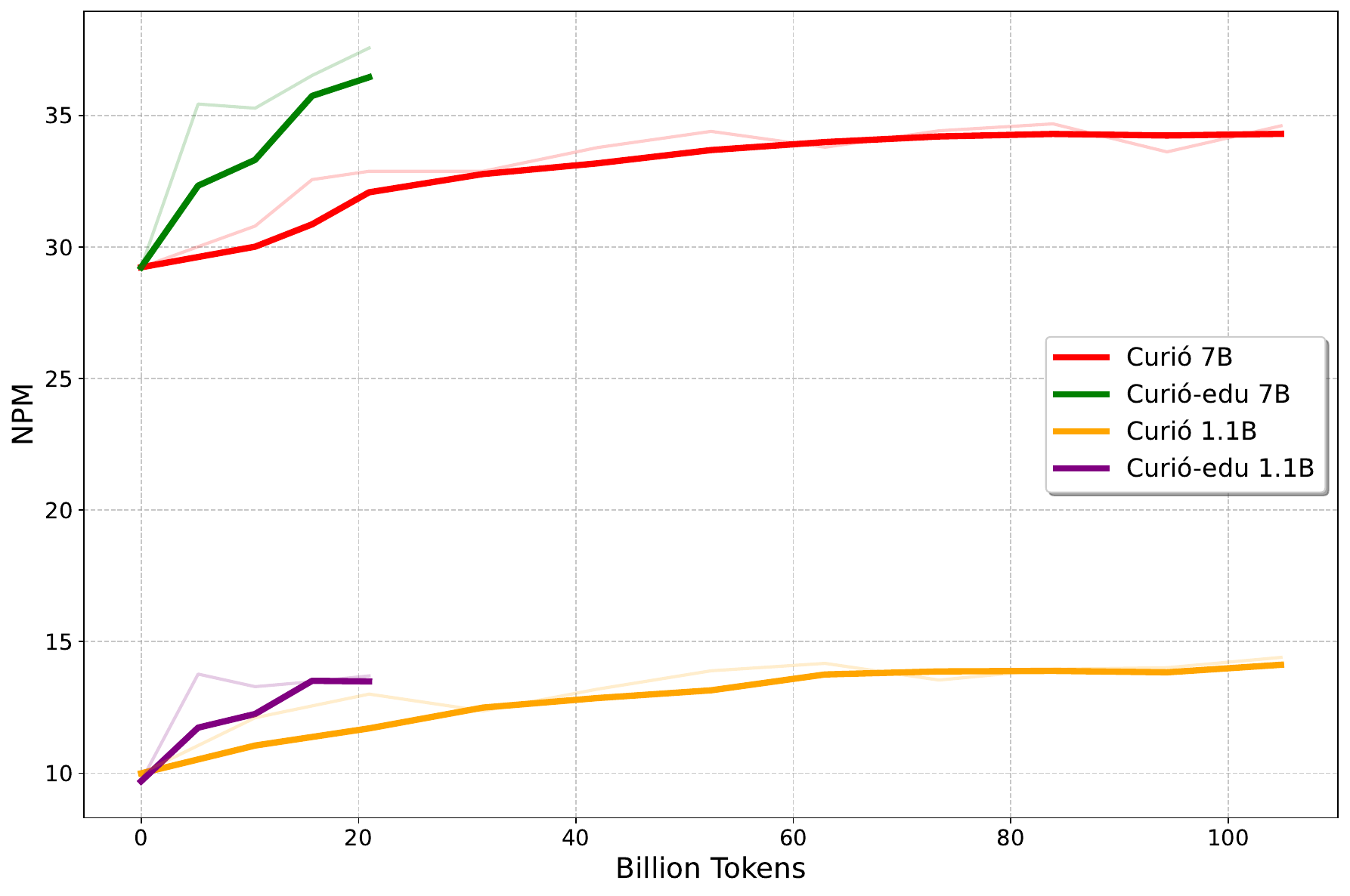}
\caption{Average NPM in PoETa~V2 for Curió~7B, Curió-Edu~7B, Curió~1.1B, and Curió-Edu~1.1B in function of the number of trained tokens. Note that the educational variants of models were trained for only 20B tokens.}
\label{fig:main_run_results}
\end{figure}

Figure~\ref{fig:main_run_results} shows the progression of overall NPM on PoETa~V2 for all models throughout training. Training on the full ClassiCC-PT corpus yields steady gains, with Curió 7B improving from an initial 29.5~NPM to a peak of 34.5 after around 80~billion tokens, after which performance plateaus with only minor oscillations until the 100~billion-token mark.

In contrast, models trained on the \textit{Education+STEM} subset achieve much faster early improvements: Curió-Edu~7B surpasses 32~NPM within the first 5~billion tokens and reaches 36.3~NPM at the end of training—roughly two points higher than the full-corpus model, despite using only one-fifth of the computational budget.

At the smaller scale, the overall trends remain similar, albeit at lower absolute performance levels. Curió~1.1B (full corpus) improves from 10.2 to 14.6~NPM, while its educational counterpart reaches 14.1~NPM after 20~billion tokens—again showing that semantically filtered data can match the gains of a much larger dataset. Unlike the 7B experiments, however, the educationally filtered variant does not surpass the full-corpus model at this scale, suggesting that targeted filtering yields its strongest benefits when model capacity is sufficient to leverage the higher-quality input.

\subsection{Performance by Subcategory}

\begin{figure*}[!htb]
\centering
\subfloat[Brazil]{\includegraphics[width=0.31\linewidth]{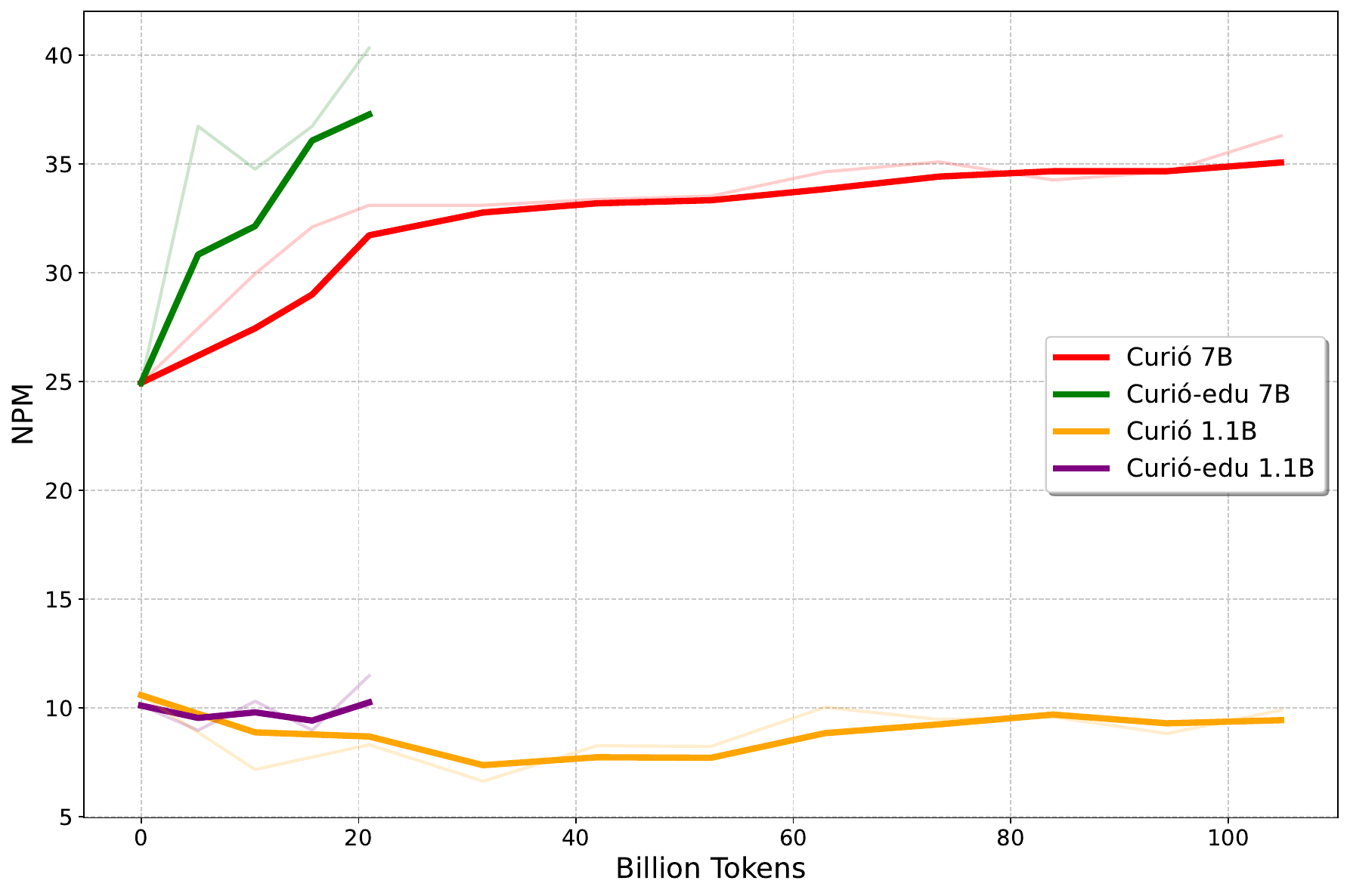}} \hspace*{0.5cm}
\subfloat[Text Understanding]{\includegraphics[width=0.31\linewidth]{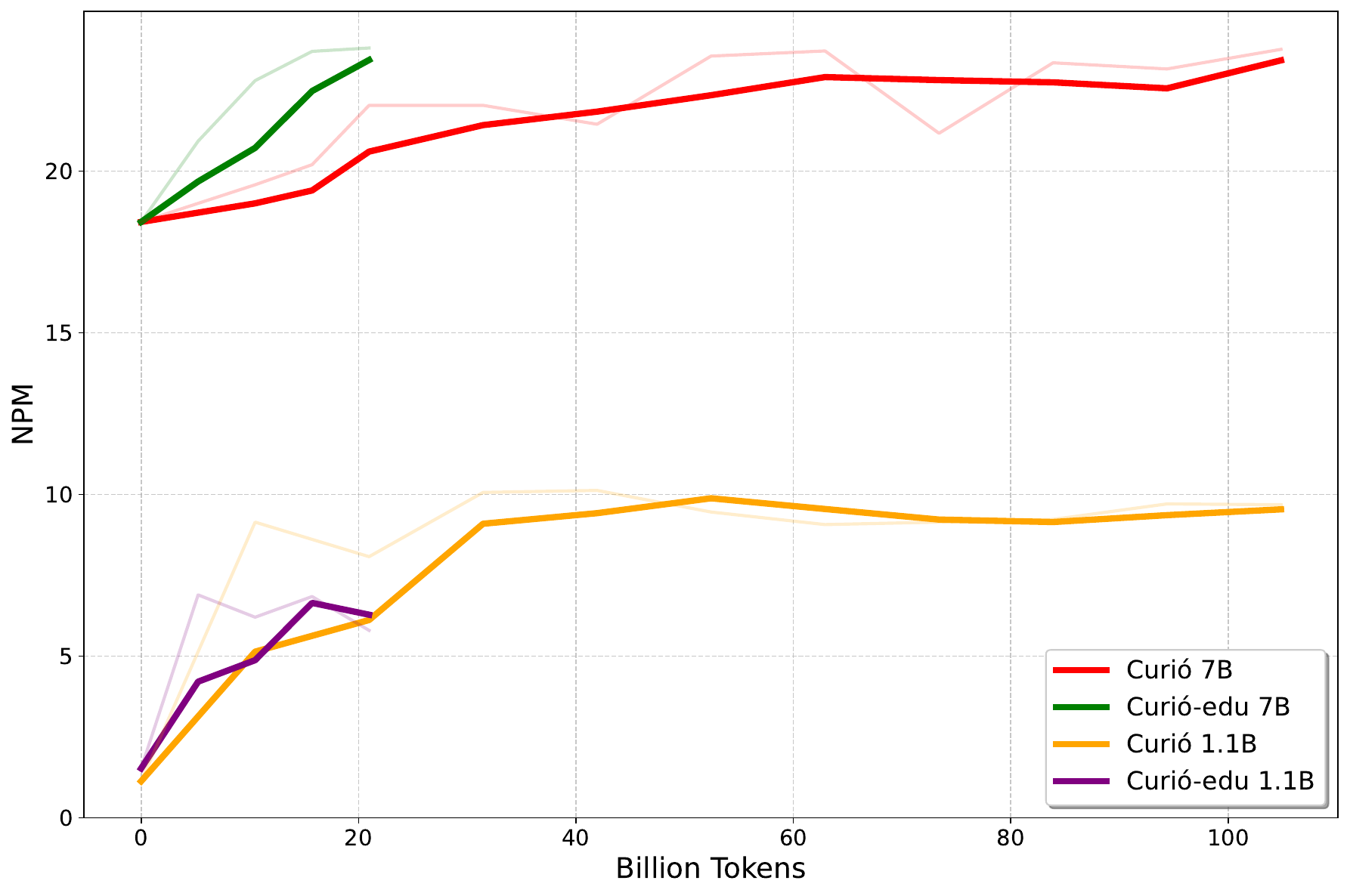}} \hspace*{0.5cm}
\subfloat[Exams]{\includegraphics[width=0.31\linewidth]{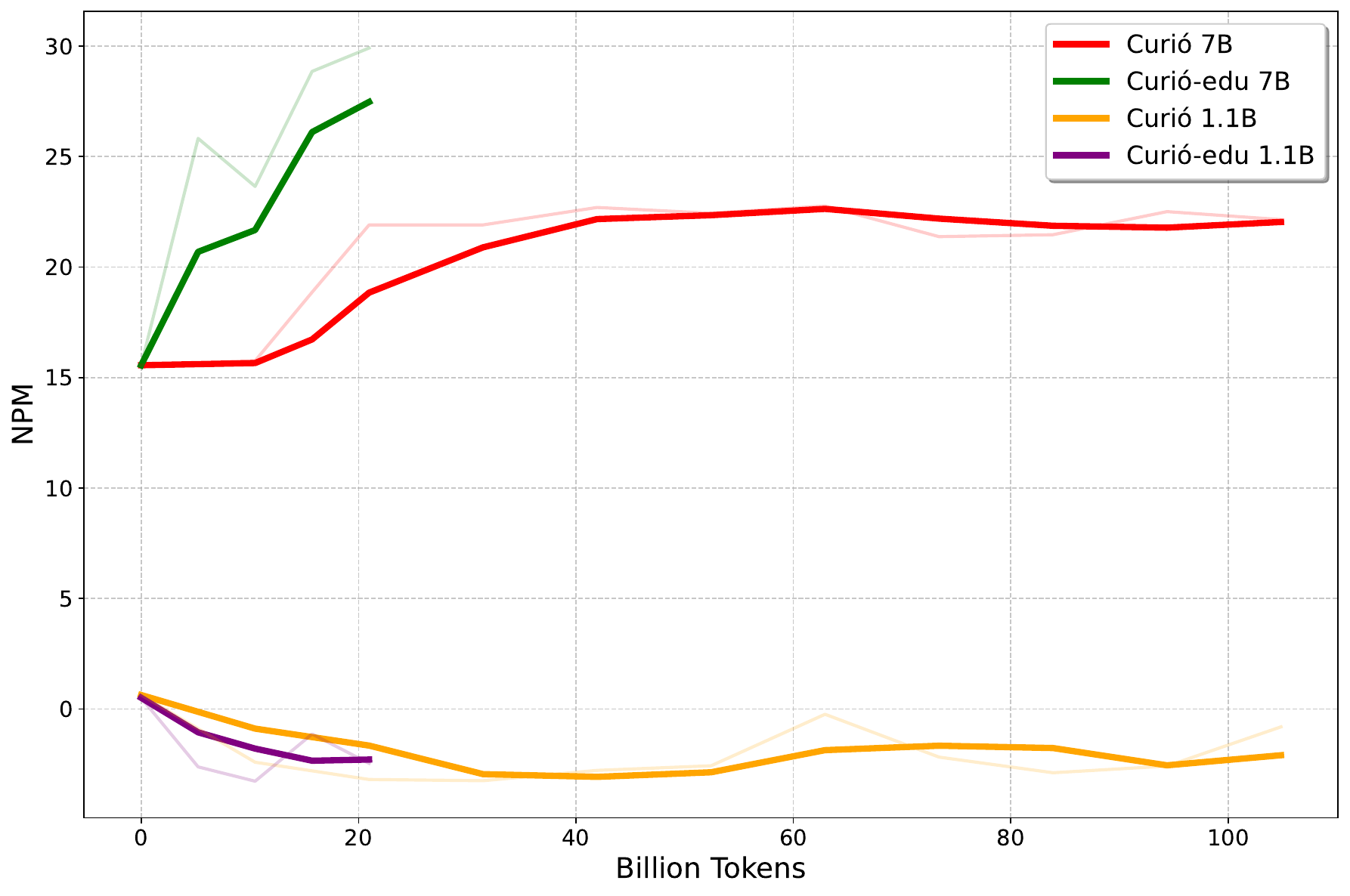}} \\   
\subfloat[Reasoning]{\includegraphics[width=0.31\linewidth]{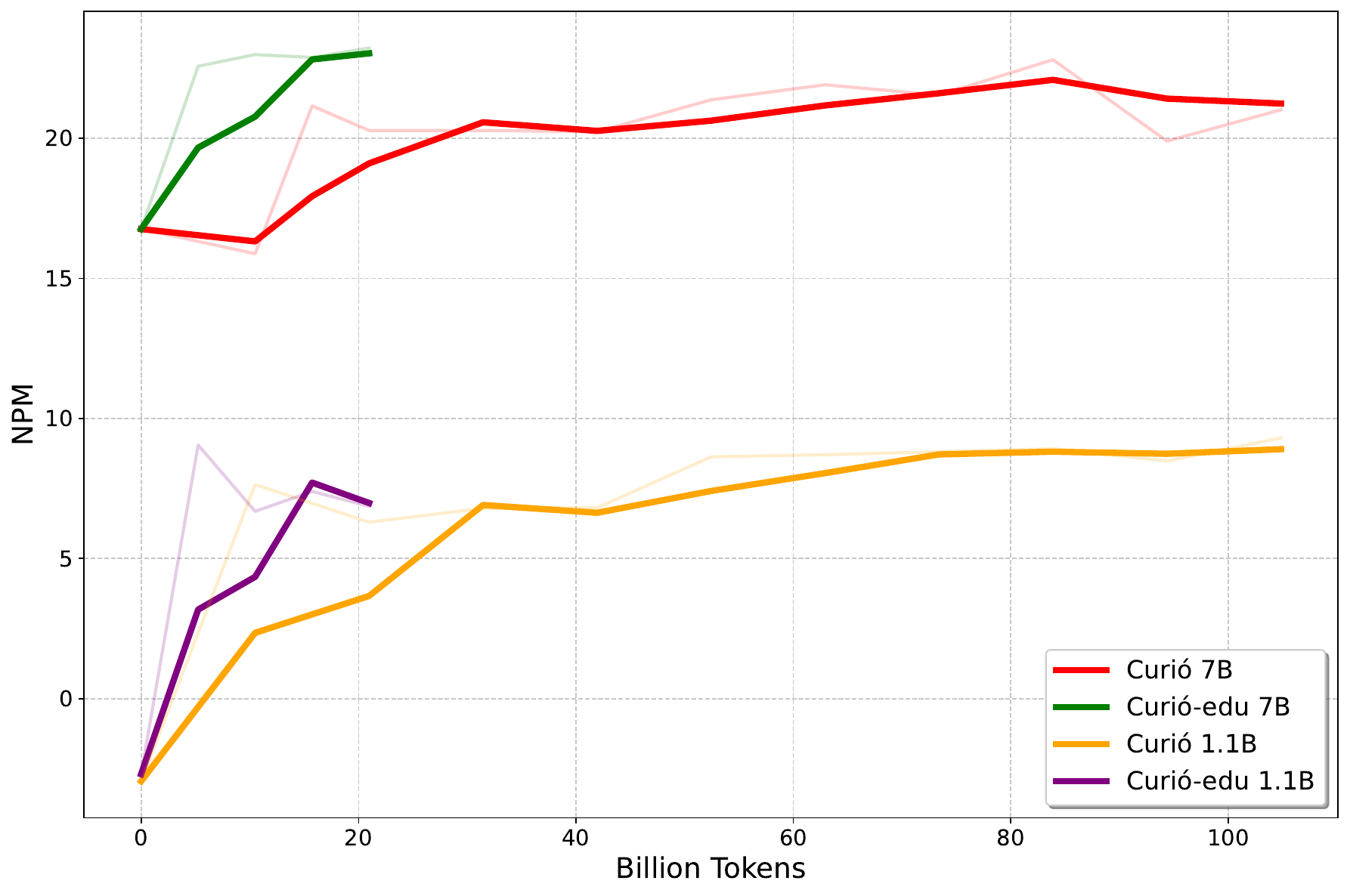}} \hspace*{0.5cm}
\subfloat[Common Sense]{\includegraphics[width=0.31\linewidth]{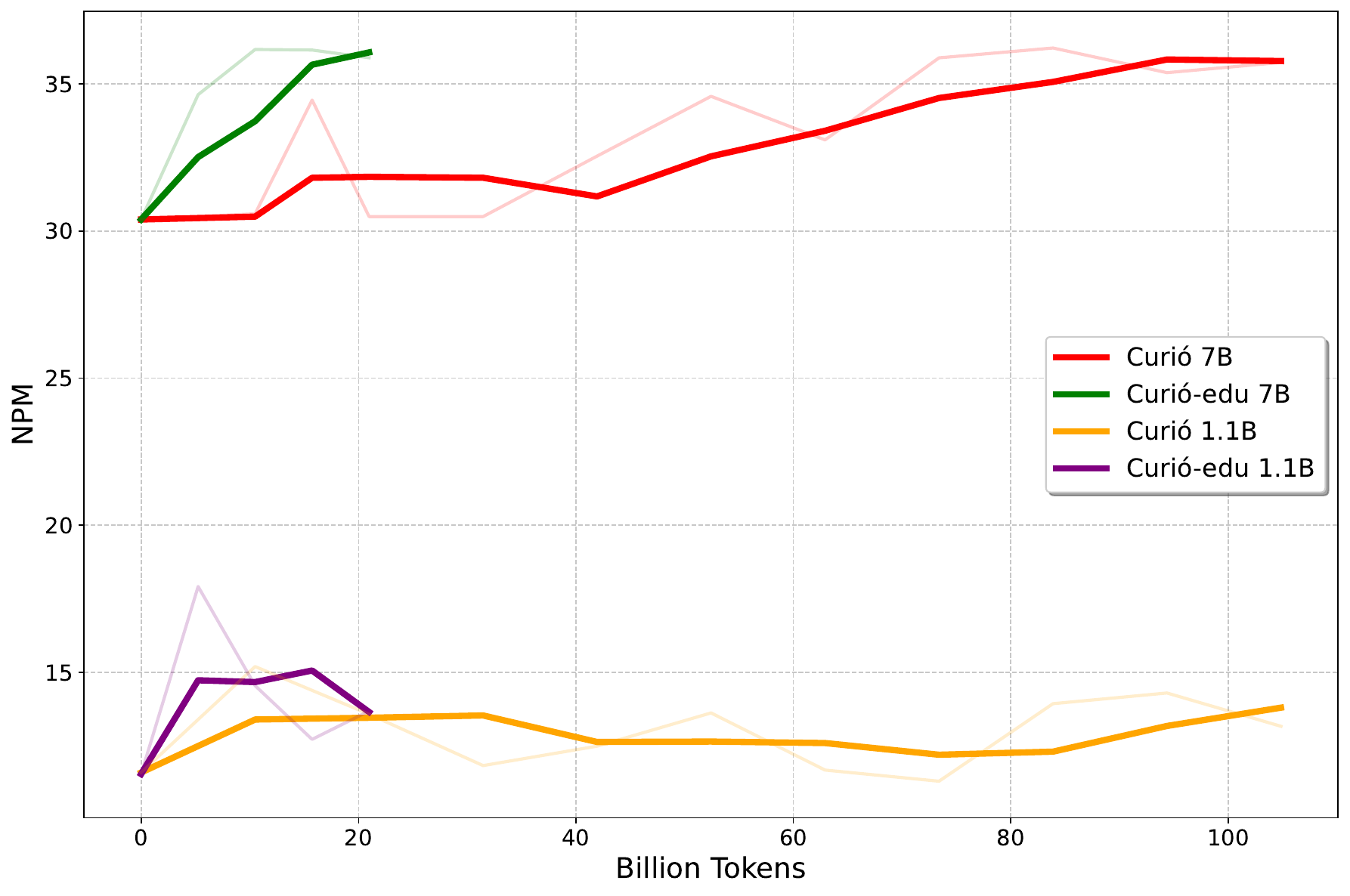}} \hspace*{0.5cm}
\subfloat[Math]{\includegraphics[width=0.31\linewidth]{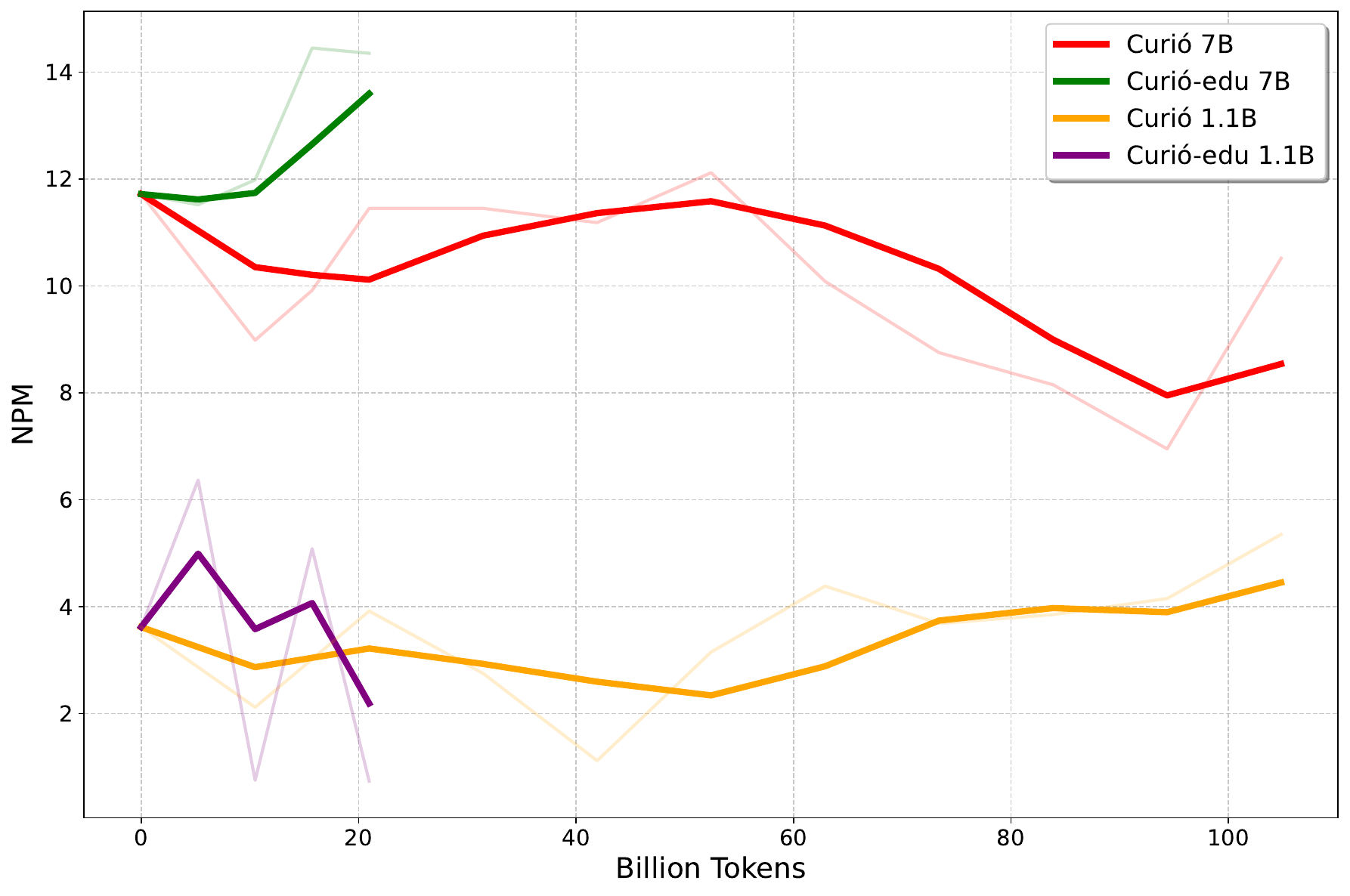}} \\
\subfloat[Social Media]{\includegraphics[width=0.31\linewidth]{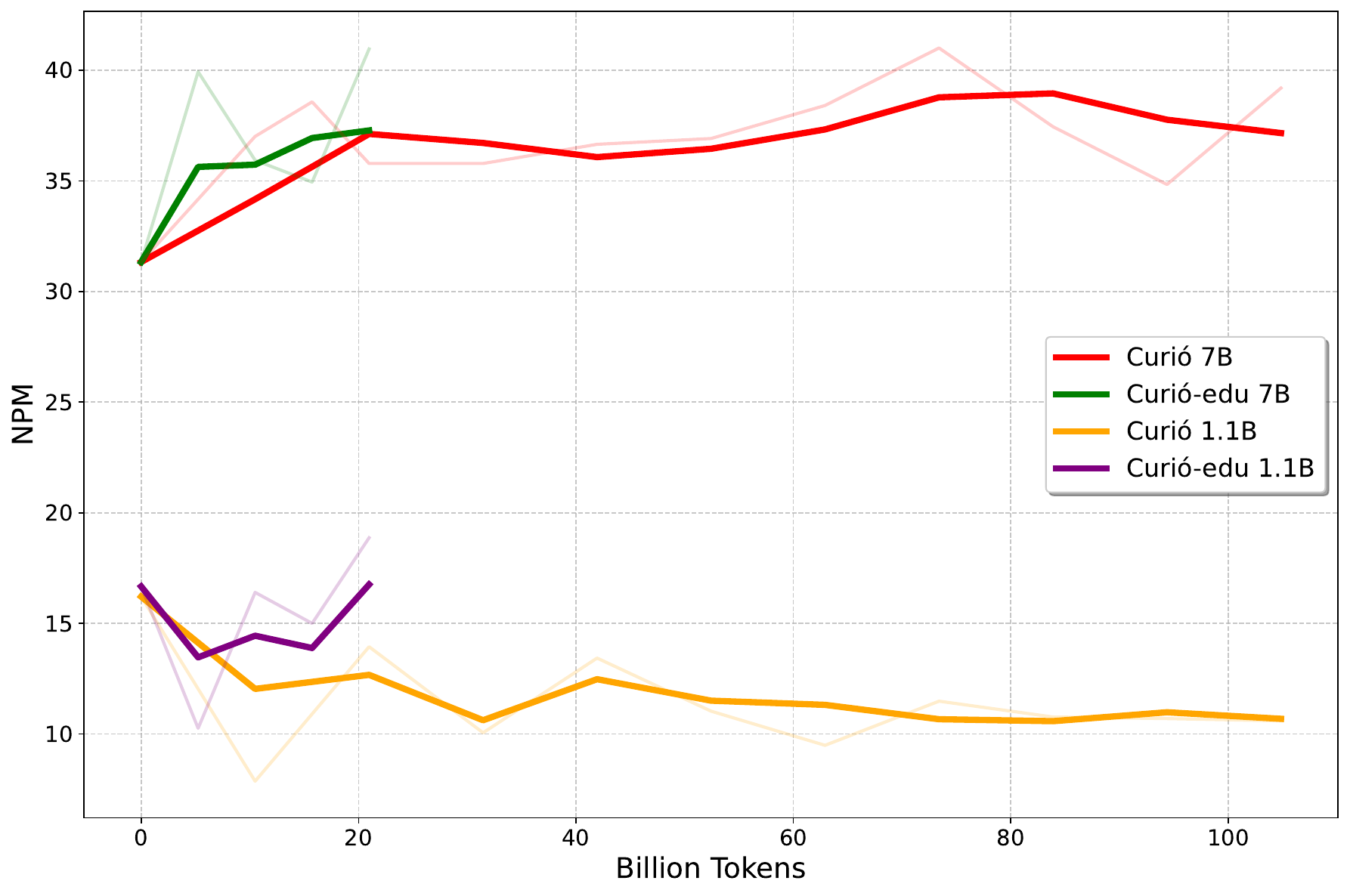}} \hspace*{0.5cm}
\subfloat[General Knowledge]{\includegraphics[width=0.31\linewidth]{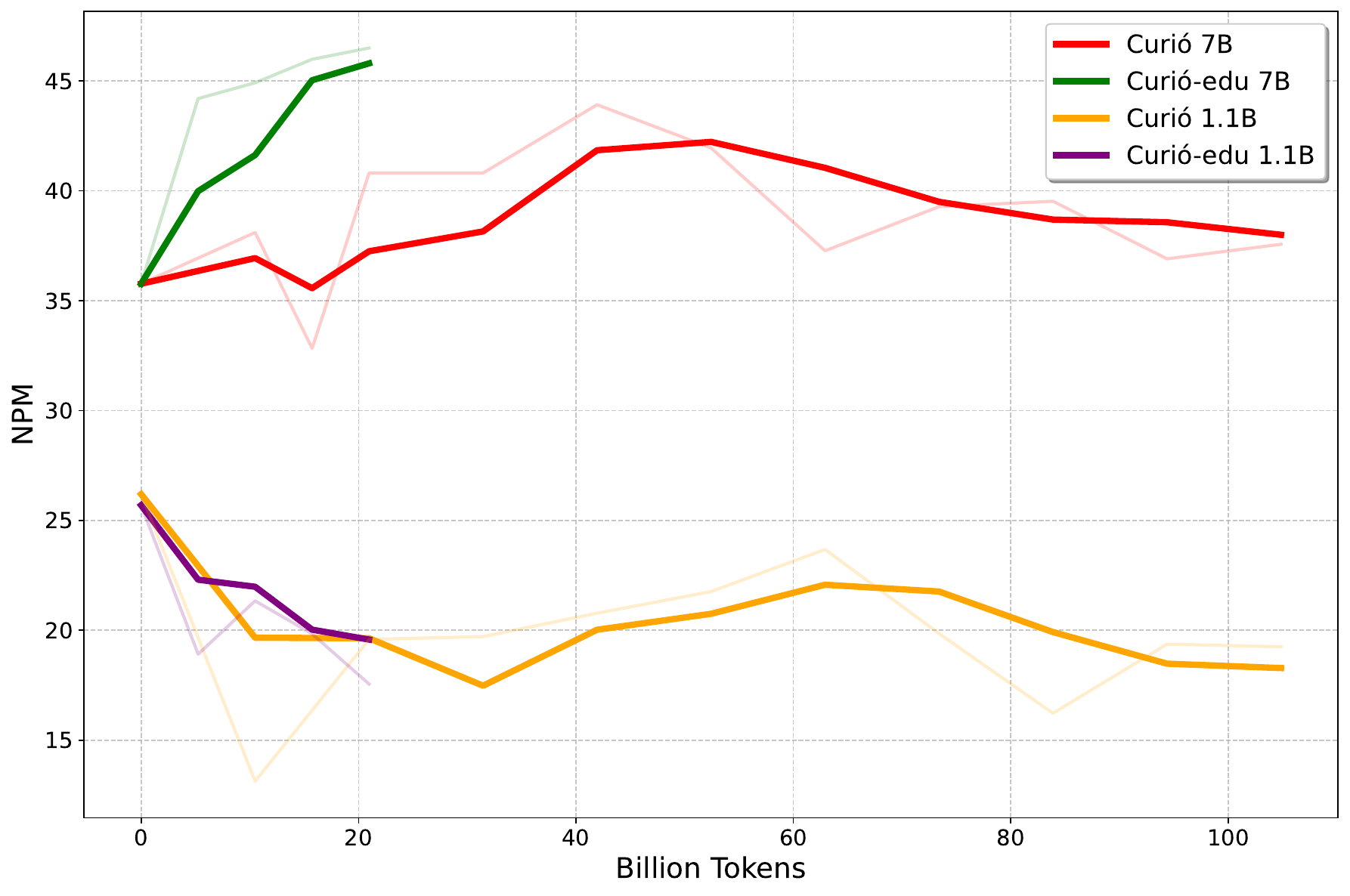}} \hspace*{0.5cm}
\subfloat[Ethics]{\includegraphics[width=0.31\linewidth]{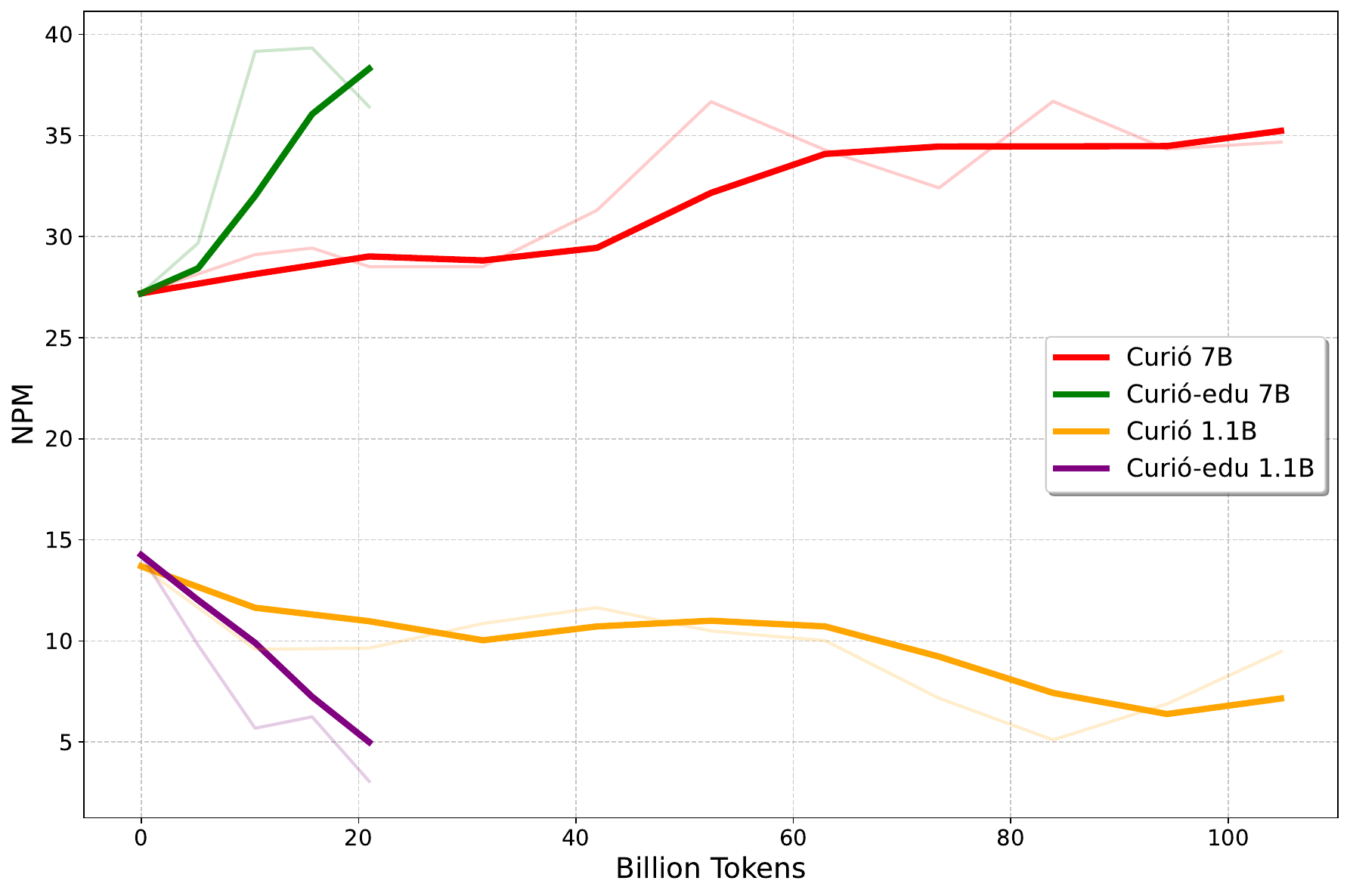}}
\caption{Average NPM across task subcategories plotted against the number of training tokens. The nine subcategories shown are those with the largest number of tasks in PoETa~V2.}
\label{fig:performance_progression_in_subcategories}
\end{figure*}

Figure~\ref{fig:performance_progression_in_subcategories} presents NPM trajectories for the nine PoETa~V2 subcategories with the largest number of tasks, providing a detailed view of how each model family evolves during continued pretraining and how data curation interacts with model scale across a diverse set of linguistic and reasoning skills.

\textbf{We observe qualitatively distinct behaviors across model scales.} In the 7B setting, the educationally filtered mixture leads to consistent and often substantial improvements across nearly all subcategories. By contrast, the 1.1B models exhibit a more uneven pattern: some domains benefit from the filtered data, others show little change, and a few even experience minor regressions. This contrast highlights a scale-dependent effect—larger models appear more capable of exploring the higher-quality, more structured signal introduced by the educational filtering, whereas smaller models may be more sensitive to the reduced data diversity and thus unable to capitalize on the same advantages.

\textbf{The 7B educational model improves across virtually every domain.} As expected, the most pronounced gains occur in \textit{Exams} and \textit{Math}, which directly align with the STEM-oriented filtering criteria. Yet improvements extend well beyond these domains: \textit{Reasoning}, \textit{Ethics}, \textit{General Knowledge}, and \textit{Common Sense} all exhibit marked increases. We hypothesize that such broad gains arise not only from the inclusion of more structured and pedagogically oriented content, but also from the removal of noisy or low-quality documents that can impede the efficiency of continued pretraining. This suggests that carefully curated data, even when topically focused, can yield generalizable benefits when model capacity is sufficient to absorb its informational structure.

\subsection{Data Impacts Under Compute- and Data-Constrained Scenarios}

We now examine two complementary training setups. The first assumes an \textbf{unlimited compute budget}—a data-constrained regime—in which models can process the full 100~billion tokens from ClassiCC-PT as well as the entire 10~billion-token educational subset. The second setup is explicitly \textbf{compute-constrained}, limiting each model to a total of 20~billion training tokens.

In this regime, the full-corpus model is trained on a random 20~billion-token sample from ClassiCC-PT, whereas the educational model sees two full passes over its 10~billion curated tokens. These two settings allow us to disentangle distinct questions: in the data-constrained scenario, we ask whether greater volume can compensate for lower data quality; in the compute-constrained scenario, we test the impact of targeted data selection when the token budget is fixed. Table~\ref{tab:data_and_compute_constrained} summarizes the results across all PoETa~V2 categories for both the unfiltered and educational mixtures under these conditions.

\begin{table}[!htb]
\setlength{\tabcolsep}{1.4mm}
\centering
\caption{Performance gain of education filtering over original models.}
\label{tab:data_and_compute_constrained}
\begin{tabular}{lcccccc}
\hline
\multicolumn{1}{l|}{}                   & \multicolumn{3}{c|}{7B Models}                     & \multicolumn{3}{c}{1.1B Models} \\ \hline
\multicolumn{1}{l|}{Category} &
  \multicolumn{1}{l}{Full} &
  \multicolumn{1}{l}{Edu} &
  \multicolumn{1}{l|}{$\Delta$ (Edu - Full)} &
  \multicolumn{1}{l}{Full} &
  \multicolumn{1}{l}{Edu} &
  \multicolumn{1}{l}{$\Delta$ (Edu - Full)} \\ \hline
\multicolumn{7}{c|}{Data Constrained}                                                                                       \\ \hline
\multicolumn{1}{l|}{All}                & 34.61 & 37.57 & \multicolumn{1}{c|}{\textbf{2.96}} & 14.4   & 13.7   & -0.7          \\
\multicolumn{1}{l|}{Reasoning}          & 21.01 & 23.22 & \multicolumn{1}{c|}{\textbf{2.21}} & 9.3    & 6.86   & -2.44         \\
\multicolumn{1}{l|}{Common Sense}       & 35.73 & 35.89 & \multicolumn{1}{c|}{\textbf{0.16}} & 13.17  & 13.69  & \textbf{0.52} \\
\multicolumn{1}{l|}{Exams}              & 22.15 & 29.91 & \multicolumn{1}{c|}{\textbf{7.76}} & -0.8   & -2.46  & -1.65         \\
\multicolumn{1}{l|}{Brazil}             & 36.3  & 40.33 & \multicolumn{1}{c|}{\textbf{4.04}} & 9.9    & 11.48  & \textbf{1.58} \\
\multicolumn{1}{l|}{Social Media}       & 39.19 & 40.96 & \multicolumn{1}{c|}{\textbf{1.76}} & 10.59  & 18.87  & \textbf{8.27} \\
\multicolumn{1}{l|}{Text Understanding} & 23.77 & 23.81 & \multicolumn{1}{c|}{\textbf{0.04}} & 9.68   & 5.79   & -3.89         \\
\multicolumn{1}{l|}{General Knowledge}  & 37.57 & 46.5  & \multicolumn{1}{c|}{\textbf{8.93}} & 19.25  & 17.55  & -1.7          \\
\multicolumn{1}{l|}{Code}               & 41.0  & 44.2  & \multicolumn{1}{c|}{\textbf{3.2}}  & 8.33   & 8.07   & -0.27         \\
\multicolumn{1}{l|}{Math}               & 10.52 & 14.35 & \multicolumn{1}{c|}{\textbf{3.83}} & 4.98   & 0.73   & -4.25         \\
\multicolumn{1}{l|}{Ethics}             & 34.68 & 36.42 & \multicolumn{1}{c|}{\textbf{1.75}} & 9.47   & 3.06   & -6.41         \\ 
\\ \hline
\multicolumn{7}{c|}{Compute Constrained}                                                                                          \\ \hline
\multicolumn{1}{l|}{All}                & 32.88 & 37.57 & \multicolumn{1}{c|}{\textbf{4.69}} & 13.01  & 13.7   & \textbf{0.69} \\
\multicolumn{1}{l|}{Reasoning}          & 20.27 & 23.22 & \multicolumn{1}{c|}{\textbf{2.94}} & 6.3    & 6.86   & \textbf{0.57} \\
\multicolumn{1}{l|}{Common Sense}       & 30.49 & 35.89 & \multicolumn{1}{c|}{\textbf{5.4}}  & 13.58  & 13.69  & \textbf{0.11} \\
\multicolumn{1}{l|}{Exams}              & 21.91 & 29.91 & \multicolumn{1}{c|}{\textbf{8.0}}  & -3.19  & -2.46  & \textbf{0.73} \\
\multicolumn{1}{l|}{Brazil}             & 33.11 & 40.33 & \multicolumn{1}{c|}{\textbf{7.23}} & 8.3    & 11.48  & \textbf{3.17} \\
\multicolumn{1}{l|}{Social Media}       & 35.78 & 40.96 & \multicolumn{1}{c|}{\textbf{5.17}} & 13.94  & 18.87  & \textbf{4.93} \\
\multicolumn{1}{l|}{Text Understanding} & 22.03 & 23.81 & \multicolumn{1}{c|}{\textbf{1.77}} & 8.07   & 5.79   & -2.28         \\
\multicolumn{1}{l|}{General Knowledge}  & 40.81 & 46.5  & \multicolumn{1}{c|}{\textbf{5.69}} & 19.58  & 17.55  & -2.03         \\
\multicolumn{1}{l|}{Code}               & 39.0  & 44.2  & \multicolumn{1}{c|}{\textbf{5.2}}  & 11.8   & 8.07   & -3.73         \\
\multicolumn{1}{l|}{Math}               & 11.45 & 14.35 & \multicolumn{1}{c|}{\textbf{2.9}}  & 4.35   & 0.73   & -3.62         \\
\multicolumn{1}{l|}{Ethics}             & 28.52 & 36.42 & \multicolumn{1}{c|}{\textbf{7.91}} & 9.65   & 3.06   & -6.58         \\
\hline
\end{tabular}
\end{table}

\textbf{For the 7B models, we observe consistent benefits from our data selection across all PoETa~V2 categories}, regardless of the training regime. Under compute-constrained conditions, every task category shows measurable improvement when using the educational subset. This result aligns with expectations, since both models process the same number of training tokens (20B), but the educational subset replaces random sampling with a semantically curated selection. More strikingly, even in the data-constrained setting—where the unfiltered model is trained for five times more tokens (100B vs.~20B)—the educationally filtered model still outperforms it across all categories. These findings indicate that thoughtful data selection can outweigh sheer data volume, even during continued pretraining of models with minimal prior exposure to the target language.

\textbf{For the 1.1B models, the benefits of data selection are less consistent.} At this smaller scale, results are mixed across both scenarios: some domains improve with educational filtering, while others show slight declines. Under compute-constrained conditions, the educational model achieves a modest advantage of 0.7~NPM points in overall score, whereas under data-constrained conditions it falls behind by roughly the same margin. Even so, the educational variant reaches performance comparable to the full-corpus model while requiring substantially less computation, as also illustrated in Figure~\ref{fig:main_run_results}.

Overall, these findings indicate that the influence of data quality is strongly tied to model capacity. Larger models are better positioned to leverage the structure and focus of curated datasets, translating higher-quality signals into consistent performance gains. In contrast, smaller models may lack the representational capacity required to fully benefit from more selective data, limiting the effectiveness of such filtering.

\subsection{Curió-Edu Performance Against Their Base Models}

\begin{figure}[!htb]
\centering
\subfloat[Average NPM of Curió-Edu~7B and Llama 2~7B in PoETa~V2 and its subcategories]{\includegraphics[width=\textwidth]{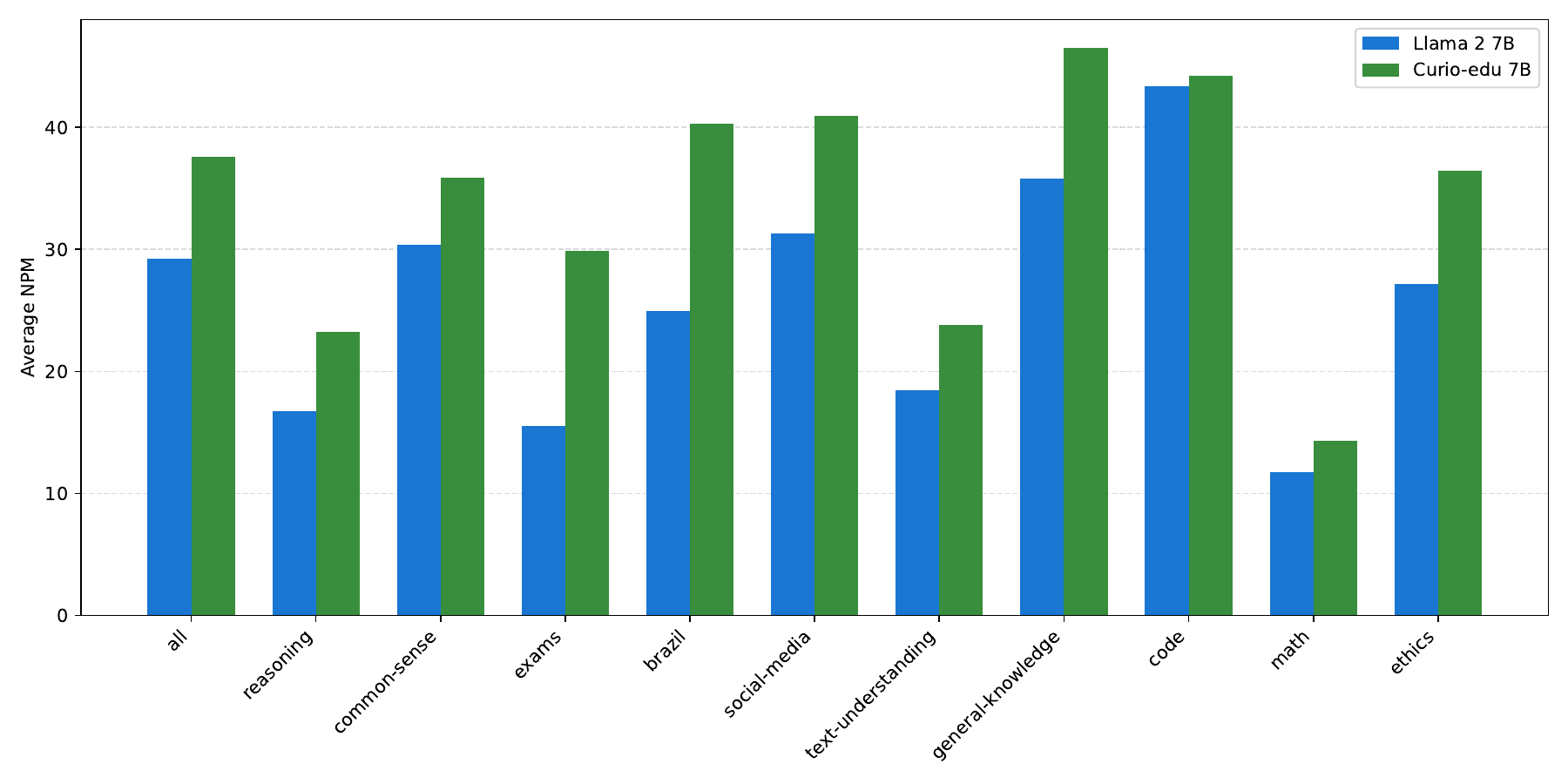} \label{fig:subfig1}} \\
\subfloat[Average NPM of Curió-Edu~1.1B and TinyLLama~1T in PoETa~V2 and its subcategories]{\includegraphics[width=\textwidth]{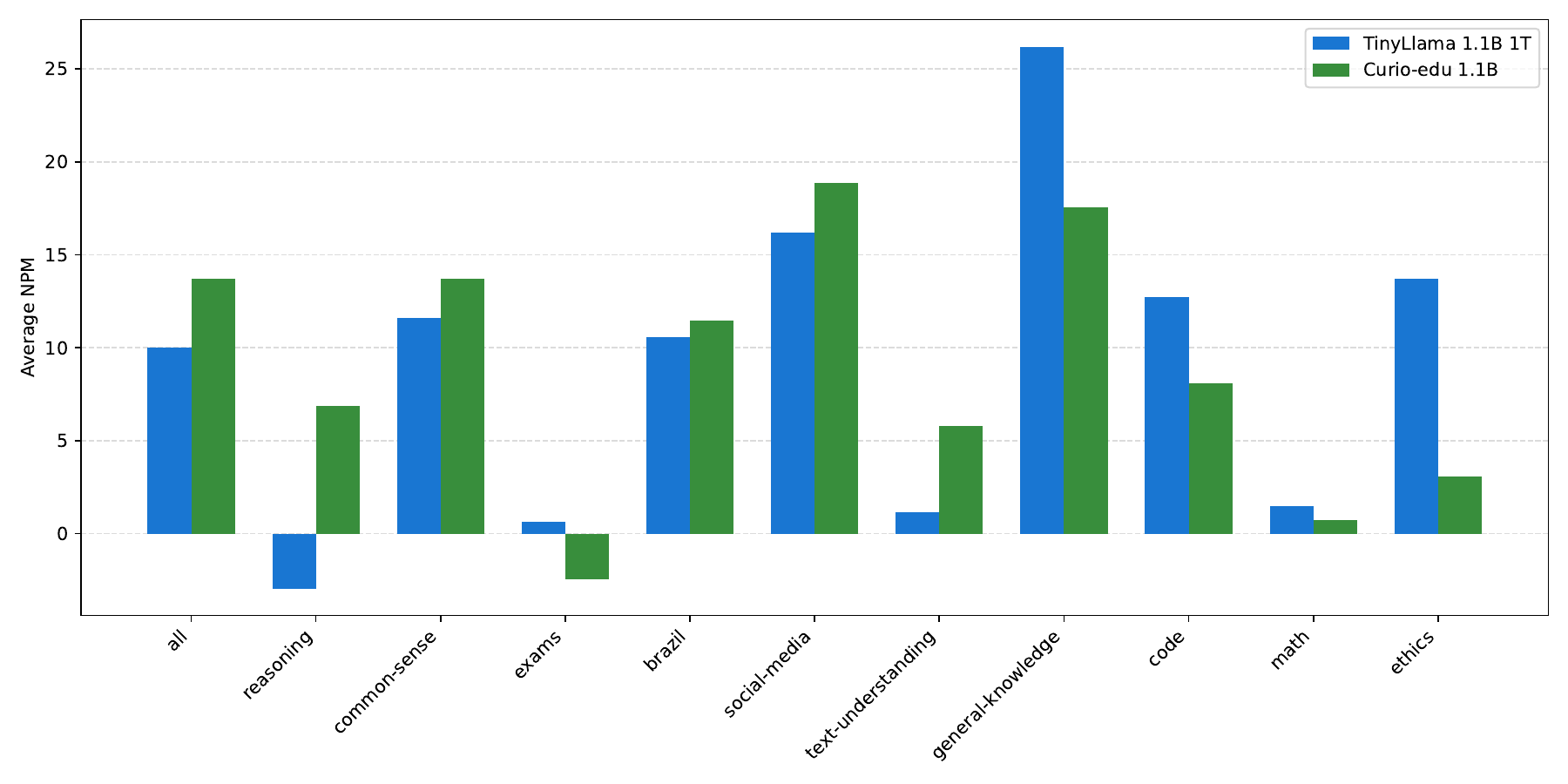} \label{fig:subfig2}}
\caption{Curió-Edu models versus their base model in PoETa~V2.}
\label{fig:curio_edu_versus_base}
\end{figure}

Figure~\ref{fig:curio_edu_versus_base} presents bar charts that provide a detailed comparison of the performance of Curió-Edu~7B and Curió-Edu~1.1B with their respective base models across the major PoETa~V2 categories, highlighting the relative gains achieved through continued pretraining on educationally filtered data.

\textbf{Curió-Edu~7B consistently outperforms LLaMA~2 across all categories of PoETa~V2.} The model attains an average NPM of 37.6 on the full benchmark, an 8.4-point improvement over the LLaMA~2~7B baseline, which scores 29.2~NPM. This corresponds to a relative gain of approximately 28\%, achieved through only 20~billion tokens of continued pretraining—representing just 0.01\% of the 2~trillion tokens used in LLaMA~2's original pretraining. These results provide compelling evidence that continued pretraining offers a cost-effective and scalable pathway for enhancing LLM performance in underrepresented languages, even when the starting point is a model with minimal prior exposure to the target language.

\textbf{Curió-Edu~1.1B also shows notable gains over its base model, though with less consistent results across categories.} On the overall PoETa~V2 benchmark, the model achieves an average NPM of 13.7, compared to 10.1 for its base model, TinyLLaMA-1T—an improvement of approximately 35\%. This increase indicates that even a relatively small model can benefit meaningfully from targeted continued pretraining on a carefully filtered corpus. However, the performance profile across subcategories reveals a more heterogeneous pattern. While several domains exhibit clear and substantial gains, others—such as \textit{Ethics}, \textit{Code}, and \textit{General Knowledge}—show modest regressions or stagnation. This uneven behavior suggests that the smaller architecture may reach its capacity limits more quickly: with fewer parameters, reduced representational power, and weaker initial coverage from the base model’s pretraining, Curió-Edu~1.1B likely struggles to internalize and generalize the educationally focused signals introduced during continued pretraining.

Overall, these comparisons reinforce the conclusion that the benefits of continued pretraining and semantic filtering scale with model capacity. Larger models appear more capable of exploiting the information present in curated datasets, effectively integrating the additional linguistic and domain-specific signals introduced during the process. In contrast, smaller models remain more sensitive to reductions in data diversity and tend to exhibit greater variability across evaluation domains, suggesting that their limited representational capacity constrains their ability to generalize from more specialized or filtered corpora.

\section{Conclusions}

In this work, we investigated how data selection and model scale shape the effectiveness of Portuguese-specific continued pretraining. Through the Curió~7B and Curió-Edu~7B models, we showed that continued pretraining on curated data can substantially improve performance even when the base model has minimal prior exposure to the target language. Using the ClassiCC-PT corpus, we demonstrated that training on only 10\% of the dataset—selected for its educational and STEM relevance—can outperform training on the full 100B-token corpus. This result highlights that data quality can outweigh sheer quantity, even in scenarios where the initial model is not well aligned with the linguistic domain of interest.

Our experiments also show that these benefits are strongly scale-dependent. Although both 7B and 1.1B models improved over their respective baselines, only the larger model consistently leveraged the advantages of semantically filtered data across all domains in PoETa~V2. This finding suggests that higher-capacity models are better equipped to exploit the structure and specificity of curated datasets, whereas smaller models may require broader and more diverse data to achieve stable generalization.

Our results reinforce continued pretraining as a practical and cost-effective strategy for enhancing LLM performance in underrepresented languages such as Portuguese. They also underscore the importance of data curation, semantic filtering, and scalable quality scoring as core components of efficient language adaptation pipelines. By combining targeted data selection with judicious use of compute, this work provides a foundation for developing more inclusive and resource-efficient large language models.

\bibliography{preprint}
\bibliographystyle{apalike}



\end{document}